\documentclass[10pt,twocolumn]{article}

\usepackage[a4paper,margin=18mm,columnsep=6mm]{geometry}
\usepackage[T1]{fontenc}
\usepackage[utf8]{inputenc}
\usepackage{amsmath,amssymb,amsfonts}
\usepackage{newtxtext,newtxmath}
\usepackage{tipa}
\DeclareUnicodeCharacter{0261}{\textipa{g}}
\usepackage{textcomp}
\usepackage{microtype}
\usepackage{booktabs}
\usepackage{tabularx}
\usepackage{array}
\usepackage{enumitem}
\usepackage{graphicx}
\usepackage[table]{xcolor}
\usepackage{tikz}
\usepackage{pgfplots}
\usepackage{caption}
\usepackage{natbib}
\usepackage{xurl}
\usepackage[colorlinks=true,allcolors=blue]{hyperref}
\usepackage[nameinlink,noabbrev]{cleveref}

\graphicspath{{figures/}}

\pgfplotsset{compat=1.18}
\usetikzlibrary{positioning,arrows.meta,calc}

\setlist[itemize]{leftmargin=*,topsep=2pt,itemsep=1pt,parsep=0pt}
\newcolumntype{Y}{>{\raggedright\arraybackslash}X}
\newcommand{\model}{BranchShine}
\newcommand{\ipacer}{\textsc{IPA-CER}}
\newcommand{\pfer}{\textsc{PFER}}
\newcommand{\params}{33.38M}
\newcommand{\code}[1]{\texttt{#1}}

\definecolor{branchblue}{RGB}{31,78,121}
\definecolor{branchlight}{RGB}{225,235,245}
\definecolor{softgray}{gray}{0.55}
\definecolor{palegray}{gray}{0.92}

\title{\vspace{-1.2em}\textbf{\model{}: Compact Raw-Audio-to-IPA Transcription with a RoPE E-Branchformer Encoder}}
\author{
Nikhil Navas$^{1}$ \quad Sergio Chevtchenko$^{1}$ \quad Talisson Damiao$^{2}$ \quad Saeed Afshar$^{1}$\\
$^{1}$International Centre for Neuromorphic Systems, Western Sydney University\\
$^{2}$Neurabuild
}

\date{}

\begin{document}
\maketitle

\begin{abstract}
Speech-to-IPA transcription is useful when the desired output is pronunciation rather than orthographic text, but competitive multilingual systems are often large and evaluation is sensitive to normalization choices. This paper presents \model{}, a \params-parameter raw-audio CTC recognizer with a lightweight convolutional front end and a 19-block RoPE E-Branchformer encoder. We find that \model{} provides a compact and competitive operating point for IPA transcription under matched normalization and scoring. On a 16,660-utterance multilingual test set covering 41 language labels, \model{} obtains 9.19\% whitespace-insensitive IPA character error rate, compared with 9.78\% for the 575.00M-parameter PhoneticXEUS baseline. A secondary child speech reading analysis shows a complementary operating profile: \model{} is more conservative on incorrect readings, while Whisper-Medium is stronger on exact acceptance of correct readings. Overall, the results indicate that a compact raw-audio-to-IPA model can approach much larger baselines on character-level IPA transcription.
\end{abstract}

\section{Introduction}

Most automatic speech recognition systems are optimized for orthographic transcripts. For language documentation, pronunciation assessment, reading support, and cross-lingual speech analysis, a closer representation of pronunciation is often more useful. The International Phonetic Alphabet (IPA) provides one such representation, but automatic speech-to-IPA evaluation is unusually sensitive to details that are easy to hide in a single leaderboard number: Unicode normalization, spaces, tokenization, exact-match criteria, and the mixture of languages in the test set.

This paper presents \model{}, a compact raw-audio model for IPA transcription. The guiding question is narrow: can a small CTC model remain close to much larger multilingual phone-recognition systems on the same held-out test set under the same normalization and scoring? The results suggest that it can. \model{} is close to PhoneticXEUS\citep{bharadwaj2026prism} on the primary character-level edit metric and is much smaller, while ZIPA\citep{zhu2025zipa} remains the strongest system in the comparison.

The paper separates four questions that should not be conflated: best overall accuracy, parameter efficiency, error composition, and transfer behavior on a child speech reading task. \Cref{sec:protocol} defines the data, model naming, and metrics before \Cref{sec:results} reports the corresponding evidence. The main observations are:
\begin{itemize}
    \item \model{} is the only model below 40M parameters in the main multilingual comparison, and it reaches 9.19\% \ipacer{} on the shared test set.
    \item \model{}'s advantage over PhoneticXEUS on \ipacer{} comes from fewer insertions and deletions, not from winning more utterances head-to-head.
    \item On the child speech reading benchmark, \model{} behaves conservatively: it is less likely to reproduce the reference transcription for incorrect recordings.
\end{itemize}

\section{Related Work}

\subsection{Universal phone recognition and speech-to-IPA}

Universal phone recognition has a long history, AlloVera and Allosaurus showed how multilingual allophone resources and a shared recognizer could support phone recognition across many languages, including low-resource settings \citep{mortensen2020allovera,li2020allosaurus}. Later work extended this line with phone inventories, articulatory features, and cross-lingual transfer \citep{li2021hierarchical,li2022phone}.

Self-supervised speech models then became a strong basis for cross-lingual phoneme recognition. Wav2Vec2Phoneme fine-tuned multilingual wav2vec 2.0 models and used articulatory mappings to support zero-shot phoneme recognition \citep{xu2022simple}. MultiIPA focused directly on speech-to-IPA transcription and showed that a smaller but cleaner IPA dataset can compete with larger weakly labeled resources \citep{taguchi2023multiipa}. Allophant added compositional phone embeddings and articulatory-attribute supervision for cross-lingual phoneme recognition \citep{glocker2023allophant}. These systems are direct prior work because they address multilingual phone or IPA output rather than only orthographic ASR.

\subsection{Recent large-scale phonetic models}

The recent baseline space is stronger than older multilingual phone-recognition work. ZIPA introduced efficient multilingual phone-recognition models trained on IPApack++, a large IPA-transcribed corpus, and includes small and large variants \citep{zhu2025zipa}. POWSM broadens the task scope by jointly handling phone recognition, ASR, grapheme-to-phoneme conversion, and phoneme-to-grapheme conversion \citep{li2025powsm}. PRiSM argues that phone recognizers should be evaluated not only by surface transcription accuracy, but also by downstream utility and more standardized probing \citep{bharadwaj2026prism}. PhoneticXEUS builds on the XEUS multilingual speech encoder and reports strong universal phone-recognition results with self-conditioned CTC over more than 100 languages \citep{chen2024xeus,bharadwaj2026phoneticxeus}.

These papers set the boundary for \model{}'s claims. ZIPA is the strongest compact-efficiency comparison in the current literature, POWSM occupies the broader phonetic foundation-model framing, and PhoneticXEUS is a strong recipe-based multilingual phone-recognition baseline. \model{} is therefore best positioned as a smaller competitive model, not as an overall state-of-the-art system.

\subsection{Compact speech encoders and low-resource transfer}

\model{} uses an E-Branchformer-style encoder. E-Branchformer was introduced for speech recognition as a way to merge local convolutional processing with global attention-like context \citep{kim2022ebranchformer}. Moonshine is related only as background on compact raw-audio ASR design and rotary position embeddings; \model{} is not presented as a Moonshine model \citep{jeffries2024moonshine}. Work on weakly phonetic supervision, including Whistle, supports the broader idea that phonetic supervision can help multilingual and low-resource transfer \citep{yusuyin2024whistle}. Dataset-quality work also matters: recent auditing studies and PRiSM-style evaluation show that phonetic labels and metric choices can strongly affect conclusions \citep{samir2025audit,bharadwaj2026prism}. This is why the child speech reading benchmark is framed here as a secondary low-resource transfer analysis under imperfect labels rather than as a broad state-of-the-art result.

\section{Evaluation Protocol}
\label{sec:protocol}

\subsection{Compared systems}

All results are reported under standardized model names. In the main tables, \model{} denotes the selected multilingual model; additional \model{} runs are not treated as separate model families.

The same convention is applied to the baselines: ZIPA-CTC-NS, ZIPA-CTC, PhoneticXEUS, POWSM-CTC, W2V2P LV-60, W2V2P XLSR-53, and MultiIPA.

\subsection{Multilingual IPA test set}

The multilingual comparison is scored on a shared test set of 16,660 utterances. The test set spans 41 language labels and 25.6 hours of audio. It is not uniformly distributed: the five largest labels - Kinyarwanda (\code{rw}, 3,233 utterances), Catalan (\code{ca}, 2,831), English (2,254), Sinhala (1,780), and Chinese (1,201) - account for 67.8\% of the utterances. This skew makes language-level generalization claims more delicate than the 41-label count alone suggests.

The corresponding training split contains 1,632,596 utterances, with 16,659 utterances used for development. Targets are IPA strings over a 112-symbol CTC vocabulary. Spaces are represented during modeling, but the primary evaluation removes whitespace because several baselines return unspaced IPA strings. The train, validation and test sets are derived from the IPApack++ dataset.

\subsection{Child speech reading benchmark}

The final transfer evaluation benchmark contains 2,340 readings labeled as correct and 2,340 labeled as incorrect. The evaluation subset contains 2,160 correct and 2,160 incorrect readings with IPA references. All reported transfer metrics use the benchmark IPA reference, keep the correct and incorrect splits separate, and report the number of evaluated items for each model. \Cref{tab:evaluation-settings} summarizes the two evaluation settings.

\begin{table}[t]
\centering
\caption{Evaluation settings. The multilingual test set is used for the primary comparison; the child speech benchmark is a secondary transfer analysis with separate acceptance and rejection criteria.}
\label{tab:evaluation-settings}
\footnotesize
\begin{tabularx}{\columnwidth}{@{}YrrY@{}}
\toprule
Setting & Train & Dev & Test / evaluated set \\
\midrule
Multilingual IPA & 1,632,596 & 16,659 & 16,660 utterances; 41 language labels; 25.6 h \\
Child speech reading & 24,010 & 1,260 & 4,680 final readings; 4,320 with IPA references \\
\bottomrule
\end{tabularx}
\end{table}

\subsection{Metrics}

Let $y_i$ be the reference IPA string and $\hat{y}_i$ be the prediction. The normalization function $N(\cdot)$ applies Unicode normalization, maps ASCII ``g'' to IPA script-g (\textipa{g}), and removes whitespace. The primary multilingual metric is a whitespace-insensitive IPA character error rate:
\begin{equation}
\mathrm{IPA\mbox{-}CER}=100\cdot\frac{\sum_i \mathrm{ED}(N(y_i),N(\hat{y}_i))}{\sum_i |N(y_i)|},
\end{equation}
where $\mathrm{ED}$ is Levenshtein edit distance. Exact match is also computed after $N(\cdot)$:
\begin{equation}
\mathrm{Exact}=100\cdot \frac{1}{n}\sum_i \mathbf{1}[N(y_i)=N(\hat{y}_i)].
\end{equation}

For the child speech reading benchmark, correct readings and incorrect readings are scored separately. On correct readings, exact match measures whether the model reproduces the reference IPA transcription. On incorrect readings, exact mismatch is used as a proxy for not accepting the expected correct transcription:
\begin{equation}
\mathrm{Mismatch}_{\mathrm{incorrect}}=100-\mathrm{Exact}_{\mathrm{incorrect}}.
\end{equation}
This metric should not be read as a complete pronunciation-assessment score; it only asks whether the model output exactly equals the reference IPA transcription for recordings marked incorrect.

A feature-aware diagnostic, \pfer{}, is also included in the analysis. It uses a PanPhon-style feature edit distance after the same basic normalization \citep{mortensen2016panphon}. Because feature mapping is less stable for some IPA outputs, \pfer{} is treated as diagnostic rather than as the primary leaderboard metric.

\section{Model}
\label{sec:model}

\begin{figure}[t]
\centering
\includegraphics[width=0.7\columnwidth]{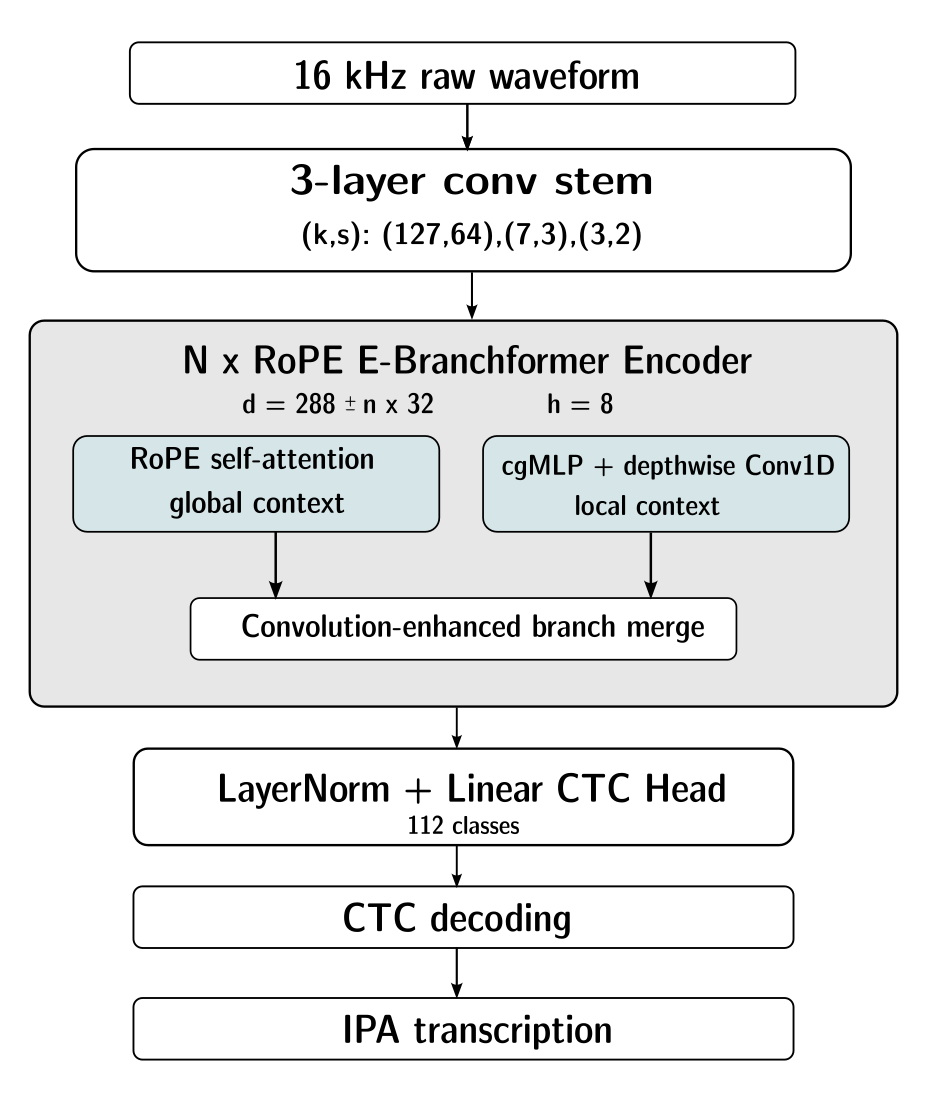}
\caption{Schematic of \model{}. The figure is intentionally high level and shows the compact raw-audio path used for IPA transcription($N,n \in \mathbb{N})$.}
\label{fig:architecture}
\end{figure}

\model{} is a raw-audio-to-IPA recognizer with a CTC objective. It uses a lightweight convolutional front end, a 19-block RoPE E-Branchformer encoder, and a linear CTC output layer over the IPA vocabulary. The encoder combines an attention branch for longer-range context with a convolutional/CGMLP branch for local acoustic structure. Rotary position embeddings provide relative position information inside attention without changing the CTC output space. \Cref{fig:architecture} gives the model path at a glance.

\Cref{tab:model-config} summarizes the model configuration. The multilingual model has 33,381,712 parameters. The child-speech adapted model has 33,382,868 parameters, a negligible increase caused by the task-specific adaptation setup. The BranchShine model, as shown in the following tables, is an order of magnitude smaller than the other models used for comparison.

\begin{table}[t]
\centering
\caption{This work uses the following \model{} configuration}
\label{tab:model-config}
\footnotesize
\begin{tabularx}{\columnwidth}{@{}lY@{}}
\toprule
Component & Value \\
\midrule
Input & 16 kHz raw waveform \\
Front end & Lightweight convolutional stem \\
Encoder & 19 RoPE E-Branchformer blocks \\
Hidden size & 288 \\
Attention heads & 8 \\
Feed-forward width & 480 \\
CGMLP width & 576 \\
CGMLP convolution kernel & 31 \\
Multilingual CTC vocabulary & 112 symbols \\
Multilingual parameters & 33,381,712 \\
Adapted-model parameters & 33,382,868 \\
\bottomrule
\end{tabularx}
\end{table}

\section{Results}
\label{sec:results}

\subsection{Main multilingual comparison}

\Cref{tab:leaderboard} reports the full multilingual comparison using one \model{} row and the baselines. ZIPA-CTC-NS is the strongest system in this comparison on both \ipacer{} and exact match. ZIPA-CTC is second. \model{} is third on \ipacer{} and is the smallest model in the table. These results should be read keeping in mind that ZIPA and PhoneticXEUS was trained on IPApack++ with a different split than the one used here, so data leakage for these two models during evaluation has not been ruled out.

The central size-aware comparison is with PhoneticXEUS. \model{} obtains 9.19\% \ipacer{} versus 9.78\% for PhoneticXEUS while using 5.8\% as many parameters. This result is meaningful but metric-dependent: PhoneticXEUS has higher exact match (20.17\% versus 18.02\%) and a better feature-aware \pfer{} diagnostic (3.21\% versus 3.92\%). Because ZIPA and PhoneticXEUS are trained in a different large-scale data regime, these results should be read as a comparison with strong reference systems rather than as a same-training-data architecture comparison.

\begin{table*}[t]
\centering
\caption{Multilingual comparison on the shared 16,660-utterance test set. Lower is better for \ipacer{} and \pfer{}; higher is better for exact match. ``Size'' is the parameter ratio relative to \model{}. The shaded row is the compact model studied here, not the best row in the table. **Training of these models on the current dataset split is required for more accurate representation of their performance on these benchmarks. Further work is being currently done in this direction.}
\label{tab:leaderboard}
\footnotesize
\begin{tabularx}{\textwidth}{@{}rYrrrrr@{}}
\toprule
Rank & Model & Params (M) & Size & \ipacer{} (\%) & Exact (\%) & \pfer{} (\%) \\
\midrule
1 & ZIPA-CTC-NS & 299.97 & 9.0$\times$ & \textbf{5.77} & \textbf{45.11} & \textbf{2.15} \\
2 & ZIPA-CTC & 299.97 & 9.0$\times$ & 6.53 & 38.40 & 2.49 \\
\rowcolor{branchlight}
3 & \model{} & 33.38 & 1.0$\times$ & 9.19 & 18.02 & 3.92 \\
4 & PhoneticXEUS & 575.00 & 17.2$\times$ & 9.78 & 20.17 & 3.21 \\
\midrule
5 & W2V2P XLSR-53** & 315.84 & 9.5$\times$ & 44.05 & 0.14 & 11.91 \\
6 & POWSM-CTC** & 425.58 & 12.8$\times$ & 45.31 & 0.08 & 12.10 \\
7 & W2V2P LV-60** & 315.84 & 9.5$\times$ & 45.50 & 0.09 & 12.24 \\
8 & MultiIPA** & 315.76 & 9.5$\times$ & 55.32 & 0.03 & 14.45 \\
\bottomrule
\end{tabularx}
\end{table*}

\begin{figure}[t]
\centering
\includegraphics[width=\columnwidth]{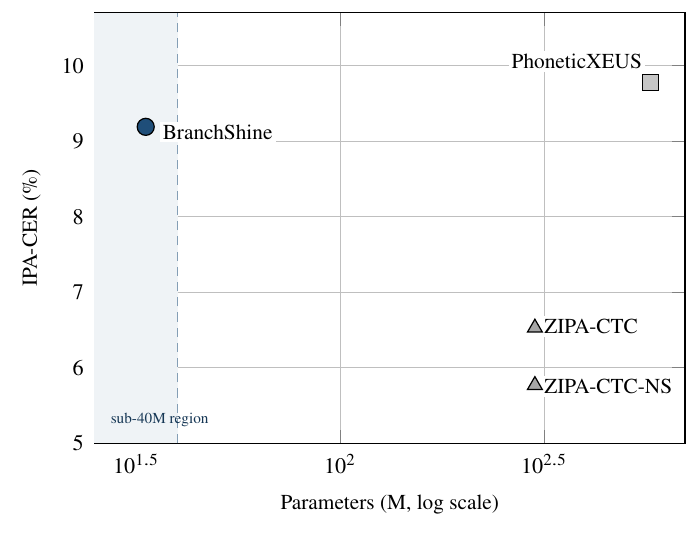}
\caption{Parameter efficiency in the main multilingual comparison. \model{} is much smaller than PhoneticXEUS and lower on \ipacer{}, while ZIPA-CTC-NS remains the best overall system.}
\label{fig:parameter_efficiency}
\end{figure}

\subsection{Comparison Explanation}

The aggregate \ipacer{} comparison does not mean \model{} is better on most utterances. \Cref{tab:h2h} shows that \model{} wins fewer non-tied utterances than PhoneticXEUS. Its aggregate advantage comes from the size of the wins: it saves 25,827 edits on utterances where it is closer to the reference, while losing 20,998 edits on utterances where PhoneticXEUS is closer, for a net saving of 4,829 edits. This explains why BranchShine achieves a lower IPA-CER even though it does not outperform PhoneticXEUS on a majority of utterances.
\begin{table}[t]
\centering
\caption{Head-to-head edit-distance comparison between \model{} and PhoneticXEUS on the multilingual test set.}
\label{tab:h2h}
\footnotesize
\begin{tabularx}{\columnwidth}{@{}Yrr@{}}
\toprule
Outcome & Utterances & Edit contribution \\
\midrule
\model{} lower edit distance & 6,289 & 25,827 fewer edits \\
\model{} higher edit distance & 6,812 & 20,998 more edits \\
Tie & 3,559 & 0 \\
\midrule
Net & 16,660 & 4,829 fewer edits \\
\bottomrule
\end{tabularx}
\end{table}

\begin{figure}[t]
\centering
\includegraphics[width=\columnwidth]{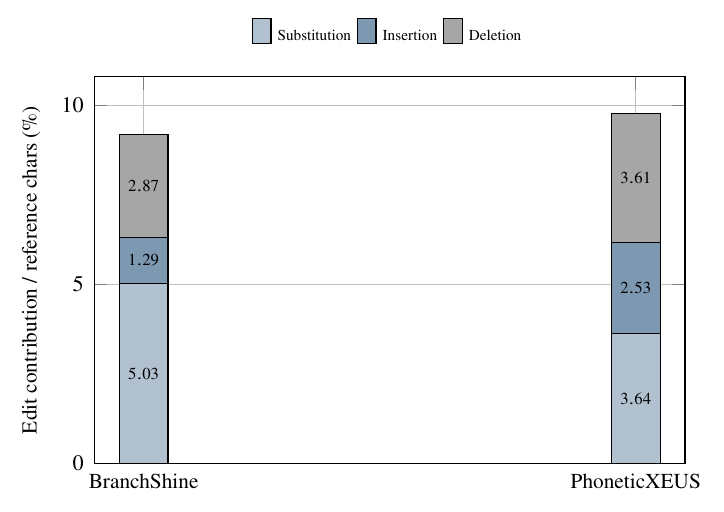}
\caption{Edit-operation decomposition for \model{} and PhoneticXEUS. \model{} has fewer insertions and deletions but more substitutions, explaining the difference between \ipacer{} and exact-match behavior.}
\label{fig:error_mix}
\end{figure}

\Cref{fig:error_mix} shows the same pattern by edit type. \model{} has more substitutions than PhoneticXEUS, but fewer insertions and deletions. This is consistent with a model that often keeps the output length closer to the reference while making more within-string symbol substitutions. That behavior can reduce character edit distance without improving strict exact match, since exact match with even a single instance of substitution is still counted as a failure in terms of scoring.

\begin{figure*}[t]
\centering
\includegraphics[width=\textwidth]{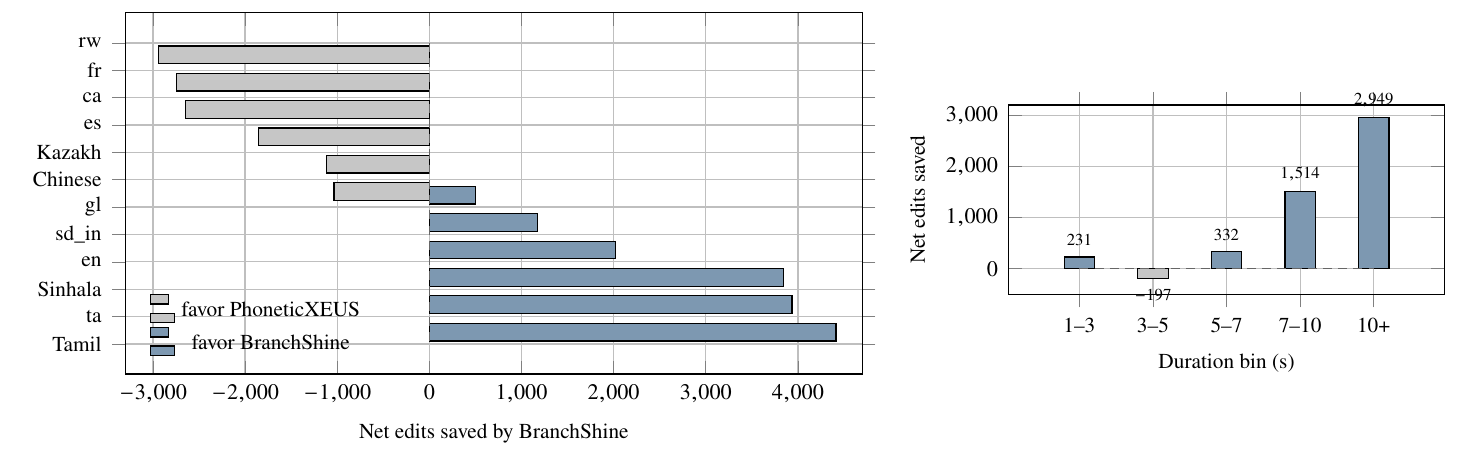}
\caption{Edit savings for \model{} relative to PhoneticXEUS, decomposed by language label and duration bin. Positive values favor \model{}; negative values favor PhoneticXEUS.}
\label{fig:edit_savings}
\end{figure*}

The aggregate edit advantage is also concentrated rather than uniform. \Cref{fig:edit_savings} decomposes the 4,829-edit net saving by language label and duration bin. Large gains on Tamil-labeled utterances, Sinhala, and English are partially offset by losses on Kinyarwanda, French, Catalan, Spanish, Kazakh, and Chinese. By duration, most of the net gain comes from utterances longer than 7 seconds, while the high-volume 3-5 second bin slightly favors PhoneticXEUS. The compactness result is therefore real, but it should not be recast as a uniform robustness claim across labels or durations.

\subsection{Ablation evidence}

\begin{figure*}[!t]
\centering
\includegraphics[width=0.8\textwidth]{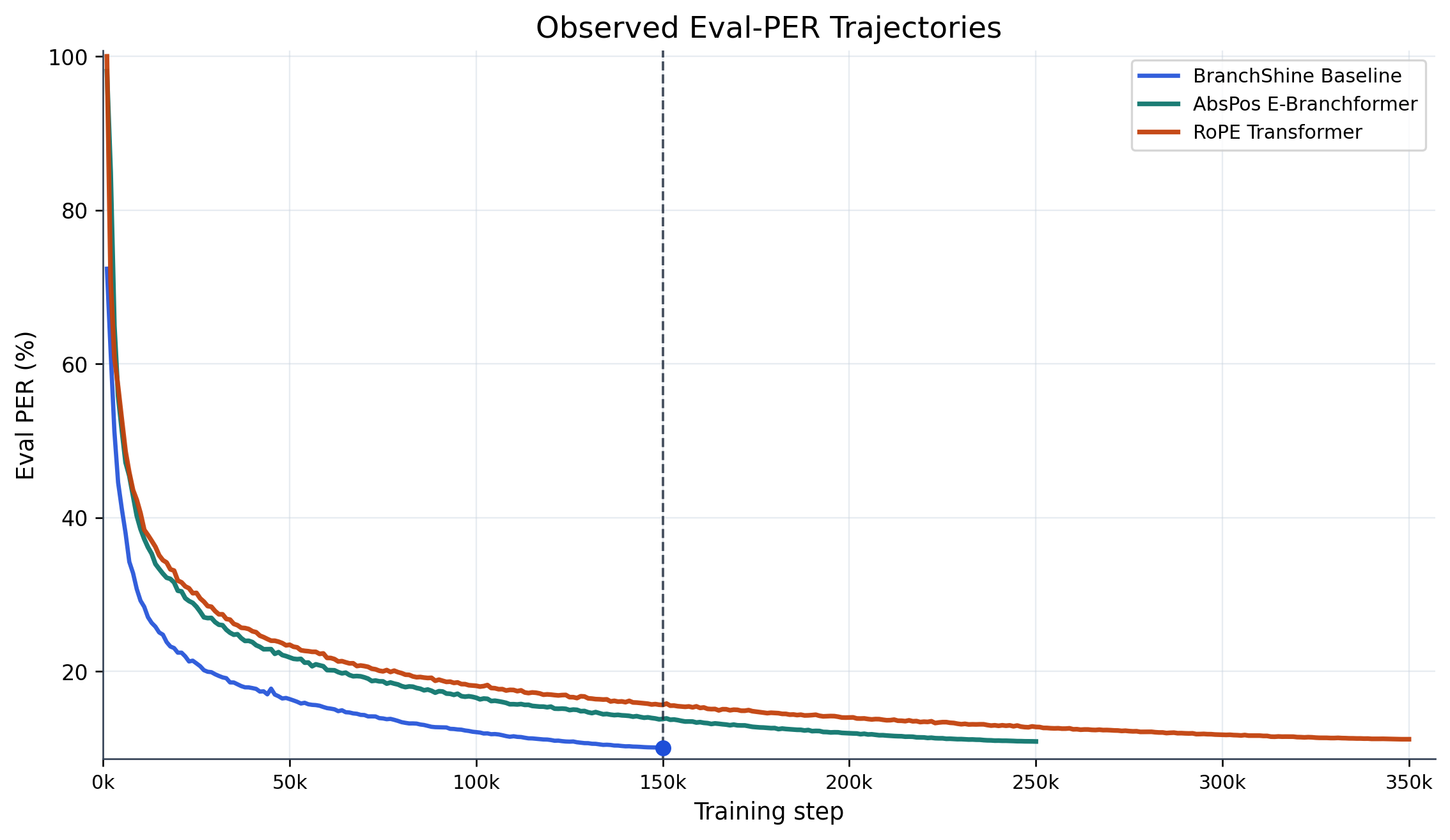}
\caption{Observed evaluation-error trajectories over the available training window for each compact variant. The dashed line marks 150k steps, where \model{} is already lower than the absolute-position E-Branchformer and RoPE Transformer are after 250k and 350k steps, respectively.}
\label{fig:ablation_trajectories}
\end{figure*}

\Cref{tab:ablation} summarizes the architecture ablation runs. These results are diagnostic rather than a complete factorial architecture study, since training histories and final-test availability differ across variants. The selected RoPE E-Branchformer configuration is best among the compared compact variants. \Cref{fig:ablation_trajectories} shows that \model{} is already lower at 150k training steps than the absolute-position E-Branchformer and RoPE Transformer are after their longer 250k- and 350k-step training windows, respectively. The absolute-position E-Branchformer is better than replacing the encoder with a RoPE Transformer, suggesting that the local/global E-Branchformer structure is important at this model size. The comparison is consistent with a benefit from RoPE on top of the E-Branchformer backbone, while leaving a fully controlled isolation of positional encoding to future work.

\begin{table}[t]
\centering
\caption{Diagnostic ablation summary. Values are best observed development \ipacer{} unless a final-test value is shown.}
\label{tab:ablation}
\footnotesize
\begin{tabularx}{\columnwidth}{@{}YrrY@{}}
\toprule
Variant & Dev & Test & Notes \\
\midrule
\model{}: RoPE E-Branchformer & 9.31 & 9.19 & Selected compact configuration \\
Absolute-position E-Branchformer & 10.88 & 10.89 & Keeps E-Branchformer, removes RoPE \\
RoPE Transformer encoder & 11.17 & 11.13 & Removes CGMLP branch and E-Branchformer merge \\
\bottomrule
\end{tabularx}
\end{table}

\subsection{Transfer to child speech readings}

\Cref{tab:child-transfer} reports selected systems on an internal child speech reading benchmark. The table intentionally reports correct readings and incorrect readings separately. A model can be conservative on incorrect readings while accepting fewer correct readings, so the two axes must be interpreted together.

Whisper-Medium has the highest correct-reading exact match among these systems at 92.55\%. \model{} is lower at 84.27\%. On incorrect readings, \model{} has 93.33\% exact mismatch, indicating a conservative operating point: it is less likely than Whisper-Medium to reproduce the reference transcription when the recording is marked incorrect. The transfer analysis is therefore an operating-point comparison: \model{} may be useful when false acceptance of incorrect readings is costly, while Whisper-Medium is preferable when correct-reading acceptance is the priority.

\begin{table*}[t]
\centering
\caption{Selected results on the internal child speech reading benchmark. Higher is better in both displayed metric columns, but the columns represent different operating goals. The evaluated counts indicate the number of items contributing to each metric.}
\label{tab:child-transfer}
\footnotesize
\setlength{\tabcolsep}{3pt}
\begin{tabularx}{\textwidth}{@{}YrrrrrY@{}}
\toprule
Model & Params (M) & Correct $n$ & Correct exact (\%) & Incorrect $n$ & Incorrect mismatch (\%) & Reading \\
\midrule
\model{} & 33.38 & 2,155 & 84.27 & 1,905 & 93.33 & Conservative rejection profile \\
HuBERT-Large & 315.49 & 2,149 & 88.88 & 1,861 & 89.09 & Balanced selected baseline \\
Whisper-Medium & 763.86 & 2,160 & 92.55 & 2,160 & 81.94 & Best correct-reading exact \\
\bottomrule
\end{tabularx}
\end{table*}

\begin{figure}[t]
\centering
\includegraphics[width=\columnwidth]{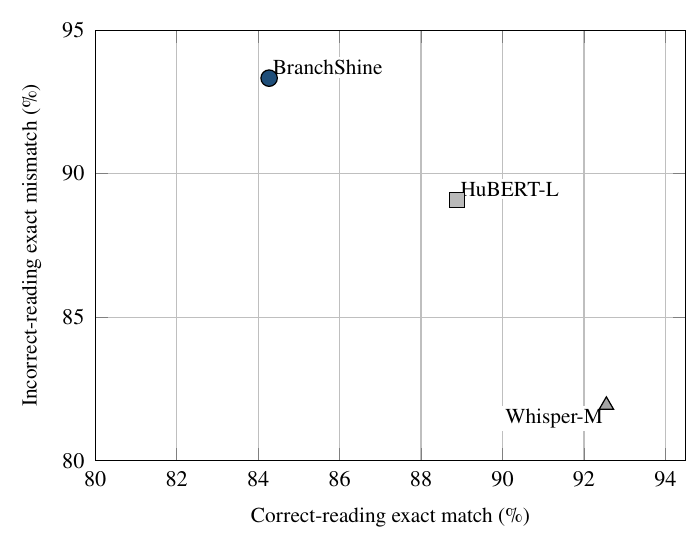}
\caption{Correct-reading exact match versus incorrect-reading mismatch on the child speech reading benchmark. \model{} is the most conservative row shown, while Whisper-Medium has the highest correct-reading exact match.}
\label{fig:child_tradeoff}
\end{figure}

\section{Discussion}

The main empirical finding is parameter-efficient competitiveness. \model{} is much smaller than PhoneticXEUS and slightly lower on the primary normalized character error metric. It also remains within a plausible performance range of contemporary universal phone-recognition systems. ZIPA-CTC-NS and ZIPA-CTC are substantially better on the same multilingual comparison, and the \pfer{} diagnostic ranks PhoneticXEUS above \model{}. The resulting picture is that \model{} is a compact and capable IPA transcription model, not a dominant model.

The diagnostic analyses clarify the aggregate metric. The head-to-head counts, edit-operation decomposition, feature-aware diagnostic, and child-speech transfer behavior all point to a metric-dependent comparison. \model{} appears especially effective at avoiding insertions and deletions, but it makes more substitutions than PhoneticXEUS. On the child speech reading benchmark, its conservative behavior on incorrect readings comes with lower correct-reading acceptance than Whisper-Medium.

These observations support a practical use case for \model{}: compact raw-audio IPA transcription where model size matters and where a character-level IPA output is sufficient for downstream screening or analysis. Questions of universal phonetic superiority, clinical diagnosis, language-family robustness, and deployment efficiency beyond parameter count remain outside the evidence reported here.

\section{Conclusion}

\model{} is a compact raw-audio-to-IPA model that performs competitively under matched normalization and scoring. \model{} reaches 9.19\% \ipacer{} on a 16,660-utterance multilingual test set while using 33.38M parameters, compared with 9.78\% for a 575.00M-parameter PhoneticXEUS baseline. The result demonstrates a strong parameter-efficiency point. At the same time, ZIPA-CTC-NS remains much better overall and PhoneticXEUS remains stronger on exact match and a feature-aware diagnostic. \model{} also performs better while considering edit saves, by demonstrating lower insertion and deletion counts than PhoneticXEUS. On the child speech reading benchmark, \model{} provides a conservative operating point rather than a universal transfer win. The main takeaway is that compact architecture design can produce capable IPA transcription, provided that model comparisons are tied to explicitly defined metrics rather than an undifferentiated leaderboard rank.

\end{document}